\documentclass{article}

\usepackage{arxiv}

\usepackage[utf8]{inputenc} 
\usepackage[T1]{fontenc}    
\usepackage{hyperref}       
\usepackage{url}            
\usepackage{booktabs}       
\usepackage{amsfonts}       
\usepackage{nicefrac}       
\usepackage{microtype}      
\usepackage{lipsum}         
\usepackage{graphicx}
\usepackage[sort&compress]{natbib}
\usepackage{doi}
\usepackage{tikz}
\usetikzlibrary{trees}

\title{Reliable Detection of Minute Targets in High-Resolution Aerial Imagery across Temporal Shifts}

\date{October 4, 2022}	

\author{ 
    Mohammad Sadegh Gholizadeh \thanks{These authors contributed equally to this work.}\\
	Shahid Rajaee University\\
	\texttt{gholizadeh@sru.ac.ir} \\
	\And
	Amir Arsalan Rezapour\footnotemark[1] \\
	Shahid Rajaee University\\
	\texttt{arsalanrzp@sru.ac.ir} \\
    \And
    Hamidreza Shayegh \\
	Shahid Rajaee University\\
	\texttt{hshayegh@sru.ac.ir} \\
    \And
    Ehsan Pazouki \\
	Shahid Rajaee University\\
	\texttt{ehsan.pazouki@sru.ac.ir} \\
}

\renewcommand{\shorttitle}{UAV Rice Seedling Detection}

\hypersetup{
pdftitle={Efficient Rice Seedling Detection in UAV Imagery using Transfer Learning with Faster R-CNN},
pdfsubject={cs.CV, cs.RO},
pdfauthor={First Author, Second Author},
pdfkeywords={UAV, Precision Agriculture, Faster R-CNN, Deep Learning},
}

\begin{document}
\maketitle

\begin{abstract}
	Efficient crop detection via Unmanned Aerial Vehicles (UAVs) is critical for scaling precision agriculture, yet it remains challenging due to the small scale of targets and environmental variability. This paper addresses the detection of rice seedlings in paddy fields by leveraging a Faster R-CNN architecture initialized via transfer learning. To overcome the specific difficulties of detecting minute objects in high-resolution aerial imagery, we curate a significant UAV dataset for training and rigorously evaluate the model’s generalization capabilities. Specifically, we validate performance across three distinct test sets acquired at different temporal intervals, thereby assessing robustness against varying imaging conditions. Our empirical results demonstrate that transfer learning not only facilitates the rapid convergence of object detection models in agricultural contexts but also yields consistent performance despite domain shifts in image acquisition.
\end{abstract}


\section{Introduction}

The escalating global demand for agricultural products places unprecedented pressure on the farming sector to transcend traditional production limitations \citep{lencucha2020government, bochtis2014advances}. As population growth necessitates higher yields, agricultural systems are increasingly compelled to adopt precision agriculture paradigms to optimize resource allocation—specifically regarding fertilizer application, irrigation management, and labor reduction—while ensuring sustainability \citep{josephson2014population, ricker2014malawi, frona2019challenge, lemouel2018land, zhang2019role, singh2020hyperspectral, yang2020rice}. In this context, the integration of high-resolution remote sensing data has become indispensable. Unmanned Aerial Vehicles (UAVs), operating within the broader Internet of Things (IoT) ecosystem, have emerged as a primary instrument for acquiring large-scale, high-fidelity field data. When coupled with advanced computer vision and deep learning techniques, these systems facilitate the transition from raw data collection to actionable ``smart farming'' insights, enabling precise interventions that address systemic production challenges \citep{tian2020computer, saiz2020smart, carolan2017publicising, lopez2017smart, zhao2021augmenting}.

While UAVs have been widely deployed for various agricultural tasks—ranging from weed classification and livestock counting to harvest estimation and aquatic product damage assessment \citep{yang2020adaptive, ward2016autonomous, yang2017spatial, driessen2014cows, li2021fast, soares2021cattle}—automated analysis remains a complex challenge. The efficacy of these systems relies heavily on the ability of computer vision algorithms to extract semantic information from complex visual data \citep{gomes2012applications, rose2018agriculture, deng2020deep, vasconez2020comparison, wu2019automatic}. This study focuses specifically on the monitoring of sowing areas, a critical early-stage phase in rice cultivation. The objective is to identify, count, and spatially locate rice seedlings to detect potential displacements or sowing failures. Unlike mature crops, rice seedlings present a unique computer vision challenge: they are minute objects with sparse feature representations, making them difficult to distinguish from background noise or visually similar categories in high-altitude aerial imagery. Traditional detection algorithms, which rely on manual feature extraction and geometric heuristics, often fail to achieve the requisite precision for such small, high-density targets \citep{yang2020semantic}.

Deep learning, particularly the use of Convolutional Neural Networks (CNNs), has fundamentally advanced the state of the art in object detection by enabling end-to-end feature learning \citep{zhang2020mask, zhang2020dense}. By training on large-scale datasets, CNN-based models can learn to robustly identify and localize targets—such as humans, vehicles, or crops—despite variations in scale and orientation \citep{kamilaris2018deep, murthy2020investigations, zou2019review, liu2020deep, etienne2021deep}. However, the detection of small objects remains a distinct sub-problem where the lack of appearance information complicates the separation of foreground from background \citep{tong2020recent}.

\begin{figure}[ht]
	\centering
	 \includegraphics[width=0.8\linewidth]{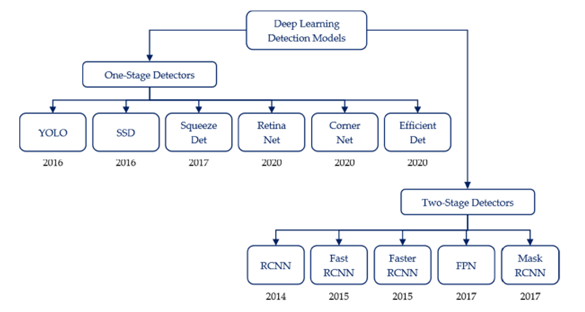} 
	\caption{Comparison of one-stage and two-stage object detection architectures. One-stage models prioritize speed, while two-stage models typically achieve higher accuracy for small objects \citep{murthy2020investigations}.}
	\label{fig:one_stage_two_stage}
\end{figure}

Current deep learning object detection architectures are generally bifurcated into one-stage and two-stage frameworks (see Figure \ref{fig:one_stage_two_stage} \citep{murthy2020investigations}).

One-stage algorithms, including YOLO \citep{redmon2016yolo}, SSD \citep{liu2016ssd}, RetinaNet \citep{lin2020focal}, and EfficientDet \citep{tan2020efficientdet}, prioritize inference speed by performing bounding box regression and class probability prediction in a single forward pass of the network. These methods extract features directly from the backbone network's output maps but often struggle with the fine-grained localization required for very small objects. Conversely, two-stage algorithms, such as R-CNN \citep{girshick2014rich}, Fast R-CNN \citep{girshick2015fast}, and Faster R-CNN \citep{ren2015faster}, introduce an intermediate step: a Region Proposal Network (RPN) that generates candidate bounding boxes (proposals) before a second stage refines these proposals for classification.

Although one-stage detectors offer computational efficiency, two-stage architectures typically yield higher Mean Average Precision (mAP) in complex scenarios involving small objects and dense clustering \citep{tong2020recent}. For instance, Mask R-CNN \citep{he2017mask} extends this capability to pixel-level segmentation. Given the specific requirements of this study—namely, the need for high localization accuracy of tiny rice seedlings rather than real-time video processing speed—we adopt the Faster R-CNN architecture.

Furthermore, the application of deep learning in agriculture is frequently hindered by the scarcity of annotated domain-specific data. To address this, we employ transfer learning, initializing our model with weights pre-trained on large benchmark datasets to accelerate convergence and improve feature extraction capabilities in a data-limited environment. This study, therefore, presents a rigorous application of a two-stage object detection framework tailored for the micro-scale detection of rice seedlings. By validating this approach on UAV-acquired imagery, we demonstrate a methodology for precision monitoring that addresses the dual challenges of small-object feature representation and agricultural data scarcity—a combination that has been insufficiently explored in traditional rice cultivation contexts.

\section{Methodology}

To facilitate early-stage growth monitoring, high-resolution aerial imagery of rice seedlings was acquired over experimental fields at the Taiwan Agriculture Research Institute, Wu-Feng District, Taichung City. Data collection utilized a DJI Phantom 4 Pro UAV equipped with a 1-inch CMOS sensor (20 MP), calibrated to capture nadir imagery across multiple temporal epochs. The dataset comprises four primary acquisition campaigns spanning 2018 and 2019. Table \ref{tab:sensor_specs} details the comprehensive sensor specifications, ensuring consistent imaging capabilities, while Table \ref{tab:flight_params} summarizes the environmental variables and flight parameters for each acquisition date \citep{yang2021uav}. The campaigns were selected to introduce specific environmental variabilities—such as lighting changes and cloud cover—to test the robustness of the detection model.

\begin{table}[ht]
	\caption{UAV imaging sensor details}
	\centering
	\begin{tabular}{ll}
		\toprule
		Parameter & Value \\
		\midrule
		Sensor Description & DJI Phantom 4 Pro \\
		Resolution & $5472 \times 3648$ pixels \\
		FOV (H$^\circ$ $\times$ V$^\circ$) & $73.7^\circ \times 53.1^\circ$ \\
		Focal Length & $8.8$ mm \\
		Sensor Size (H $\times$ V) & $13.2 \times 8.8$ mm \\
		Pixel Size & $2.41 \times 2.41$ $\mu$m \\
		Image Format & JPG \\
		Dynamic Range & 8 bit \\
		\bottomrule
	\end{tabular}
	\label{tab:sensor_specs}
\end{table}

\begin{table}[ht]
    \caption{Details of flight mission \citep{yang2021uav}}
    \centering
    \begin{tabular}{lcccc}
        \toprule
        \textbf{Parameter} & \textbf{07 Aug 2018} & \textbf{14 Aug 2018} & \textbf{12 Aug 2019} & \textbf{20 Aug 2019} \\
        \midrule
        Time & 07:19–07:32 & 07:03–07:13 & 14:23–14:44 & 08:16–08:36 \\
        Weather & Mostly clear & Mostly cloudy & Cloudy/Rain & Partly cloudy \\
        Avg. Temp ($^\circ$C) & 28.9 & 26.8 & 26.6 & 27.5 \\
        Avg. Press (hPa) & 997.7 & 992.0 & 994.1 & 996.4 \\
        Flight Height (m) & 21.4 & 20.8 & 18.6 & 19.1 \\
        GSD (mm/px) & 5.24 & 5.09 & 4.62 & 4.78 \\
        Fwd Overlap (\%) & 80 & 80 & 85 & 85 \\
        Side Overlap (\%) & 75 & 75 & 80 & 80 \\
        Collected Images & 349 & 299 & 615 & 596 \\
        \bottomrule
    \end{tabular}
    \label{tab:flight_params}
\end{table}

Following acquisition, raw imagery was orthorectified and mosaicked to generate a unified field map. To satisfy GPU memory constraints and maximize the effective receptive field relative to object size, the high-resolution mosaics were tessellated into sub-images of dimension $512 \times 512$ pixels. Ground truth annotation was performed using the labelImg tool, where agricultural experts manually bounded individual rice seedlings. Given the high labor cost of pixel-level annotation, a semi-automated proposal method was employed to accelerate the process, followed by manual verification. The resulting dataset was partitioned into a training set containing 273 sub-images and a validation set of 60 sub-images, both derived from the 07 Aug 2018 campaign. To evaluate model robustness against temporal domain shifts, three distinct generalization test sets were curated from the subsequent dates (14 Aug 2018, 12 Aug 2019, 20 Aug 2019).

\begin{figure}[ht]
	\centering
	\includegraphics[width=0.8\linewidth]{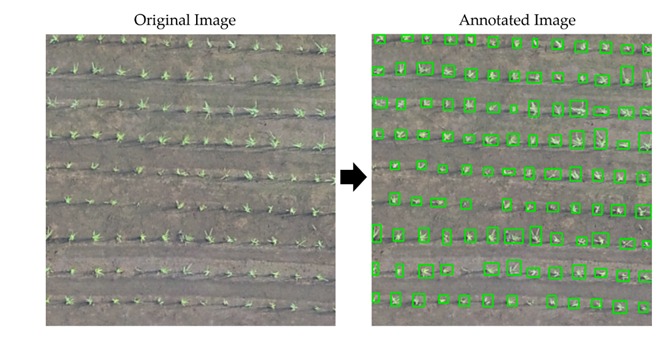}
	\caption{Rice seedlings annotated with bounding boxes in green on sub-images}
	\label{fig:study_area}
\end{figure}

\begin{figure}[ht]
	\centering
	\includegraphics[width=0.8\linewidth]{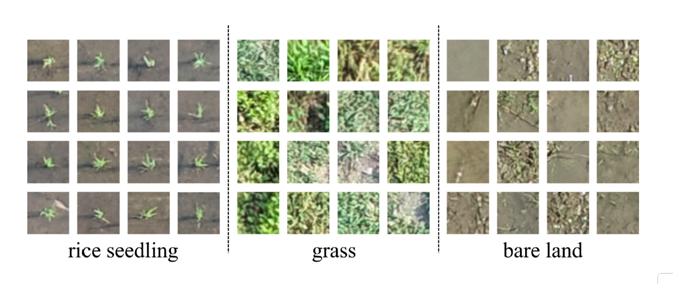}
	\caption{Examples of annotated images for model training.}
	\label{fig:study_area}
\end{figure}

\begin{figure}[ht]
	\centering
	\includegraphics[width=0.8\linewidth]{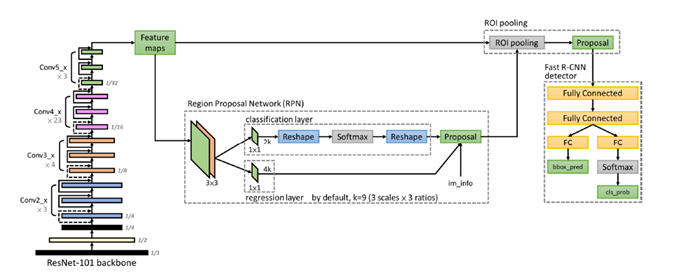}
	\caption{ Architecture of Faster R-CNN ResNet-101.}
	\label{fig:study_area}
\end{figure}

For the detection task, we employ the Faster R-CNN architecture \citep{ren2015faster}, a two-stage object detector that integrates a Region Proposal Network (RPN) with a Fast R-CNN detector. This architecture is particularly well-suited for high-precision localization of small, dense targets where single-stage detectors often degrade. The network utilizes a ResNet-101 backbone \citep{he2016deep}, pre-trained on the COCO dataset to leverage transfer learning. The RPN operates by sliding a small window over the feature map extracted by the backbone, predicting multiple region proposals at each location based on $k$ fixed-ratio anchor boxes. The loss function for the RPN is defined as:

\begin{equation}
	L(\{p_i\}, \{t_i\}) = \frac{1}{N_{cls}} \sum_i L_{cls}(p_i, p_i^*) + \lambda \frac{1}{N_{reg}} \sum_i p_i^* L_{reg}(t_i, t_i^*)
\end{equation}

where $i$ is the index of an anchor, $p_i$ is the predicted probability of being an object, $p_i^*$ is the ground truth label, $t_i$ is the vector of bounding box coordinates, and $L_{reg}$ is the robust smooth $L_1$ loss. Proposals from the RPN are subsequently passed to a Region of Interest (RoI) pooling layer for classification and bounding box refinement. The model was implemented using the TensorFlow Object Detection API. Input images were resized to $640 \times 640$ pixels during training, with a batch size of 8 and a total of 25,000 steps. The maximum number of detections per image was set to 200 to accommodate the high density of seedlings, and optimization was performed using a Momentum Optimizer with a learning rate schedule adapted for fine-tuning.

\section{Results and Discussion}

All experiments were conducted in a containerized TensorFlow 2.5 environment hosted on a workstation equipped with an Intel Xeon Gold 6154 CPU (90 GB RAM) and a single NVIDIA Tesla V100 GPU. The Faster R-CNN model was initialized with COCO 2017 pre-trained weights to accelerate convergence and leverage robust feature priors. Inference efficiency is a critical constraint for agricultural deployment; as detailed in Table \ref{tab:performance_cost}, the system achieves an average inference throughput of 20 frames per second (FPS), corresponding to a total latency of $\tau \approx 50$ ms per image. While experimental I/O was serialized from disk, deployment architectures utilizing direct bus transfer from camera cache would likely yield even lower latency.

\begin{table}[ht]
    \centering
    \caption{Model performance and computational cost}
    \label{tab:performance_cost}
    \setlength{\tabcolsep}{6pt}
    \renewcommand{\arraystretch}{1.2}

    \begin{tabular}{ccccc cccc}
        \toprule
        \multicolumn{2}{c}{\textbf{Training}} &
        \multicolumn{2}{c}{\textbf{Test}} &
        \multicolumn{4}{c}{\textbf{Computational Cost (s)}} &
        \textbf{Total (fps)} \\

        \cmidrule(r){1-2} \cmidrule(r){3-4} \cmidrule(lr){5-8}

        mAP & mIoU & mAP & mIoU &
        Preprocess & Inference & Visualization & Total & fps \\

        \midrule
        1.000 & 0.996 & 0.888 & 0.637 &
        0.005 & 0.042 & 0.003 & 0.050 & 20.000 \\

        \bottomrule
    \end{tabular}
\end{table}

Regarding detection accuracy, the model demonstrates high efficacy on the primary test set, achieving a Mean Average Precision (mAP) of $0.888$1. We observe a divergence between training performance ($mAP \approx 1.0$) and testing performance, indicating a degree of overfitting common in deep learning applications with limited domain-specific data2. However, the use of transfer learning mitigated the need for extensive training epochs (convergence $< 500$ epochs, $< 1$ hour), significantly reducing the computational budget compared to training from scratch3. Comparatively, prior work by \citet{wu2019automatic} utilized a fully convolutional architecture for rice seedling counting, reporting a correlation coefficient of $R^2 = 0.94$4. However, their approach was limited to regression without explicit localization5. By employing a two-stage detector, our method matches counting capabilities while providing precise bounding box regression and size estimation6.To ensure the model is robust enough for real-world application, we evaluated its generalization capabilities across three additional datasets acquired on different dates (14 Aug 2018, 12 Aug 2019, 20 Aug 2019)7. Table \ref{tab:generalization_metrics} summarizes the comprehensive performance metrics, including Average Precision (AP), Intersection over Union (IoU), F1-Score, Precision, and Recall across these temporal domains.

\begin{table}[ht]
    \centering
    \caption{Evaluation of Generalization across Temporal Domain Shifts. The model is trained on August 7 data and evaluated on subsequent dates to test robustness against environmental variations.}
    \label{tab:generalization_metrics}
    \setlength{\tabcolsep}{5pt} 
    \renewcommand{\arraystretch}{1.2} 
    \begin{tabular}{lcccccc}
        \toprule
        \textbf{Test Date} & \textbf{Conditions} & \textbf{AP} & \textbf{IoU} & \textbf{Precision} & \textbf{Recall} & \textbf{F1-Score} \\
        \midrule
        7 August 2018 & Clear (Ref)   & 0.888 & 0.637 & 0.855 & 0.780 & 0.783 \\
        14 August 2018 & Cloudy        & \textbf{0.981} & 0.686 & 0.948 & 0.747 & \textbf{0.904} \\
        12 August 2019 & Rain/Cloudy   & \textbf{0.986} & \textbf{0.871} & \textbf{0.972} & 0.615 & 0.790 \\
        20 August 2019 & Algae/Ponding & 0.739 & 0.382 & 0.583 & 0.345 & 0.480 \\
        \bottomrule
    \end{tabular}
\end{table}

\begin{figure}[ht]
	\centering
	\includegraphics[width=0.7\linewidth]{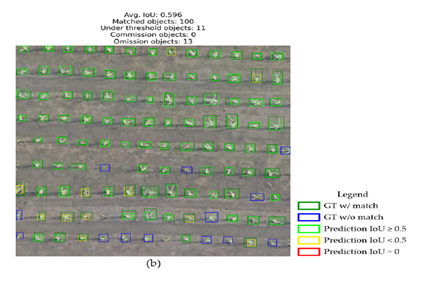} 
	\caption{Example of visualized detection results of Faster R-CNN.}
	\label{fig:generalization_analysis}
\end{figure}

The results indicate that the model generalizes effectively to the August 14 and August 12 datasets, even exceeding the baseline metrics in some instances. However, performance degrades on the August 20, 2019 dataset, where the F1-Score drops to $0.480$. Qualitative analysis (see Figure \ref{fig:generalization_analysis}) reveals that this performance drop is attributable to severe background occlusion caused by algae proliferation and varying water levels. This highlights the sensitivity of CNNs to domain shifts in texture and color, suggesting that future work must incorporate data augmentation techniques targeting environmental noise to enhance robustness.

\begin{figure}[ht]
	\centering
	\includegraphics[width=0.8\linewidth]{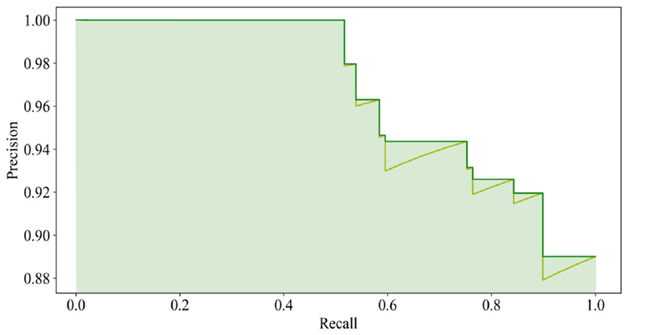} 
	\caption{AP curves of the detection results of Faster}
	\label{fig:generalization_analysis}
\end{figure}

\begin{figure}[ht]
	\centering
	\includegraphics[width=0.5\linewidth]{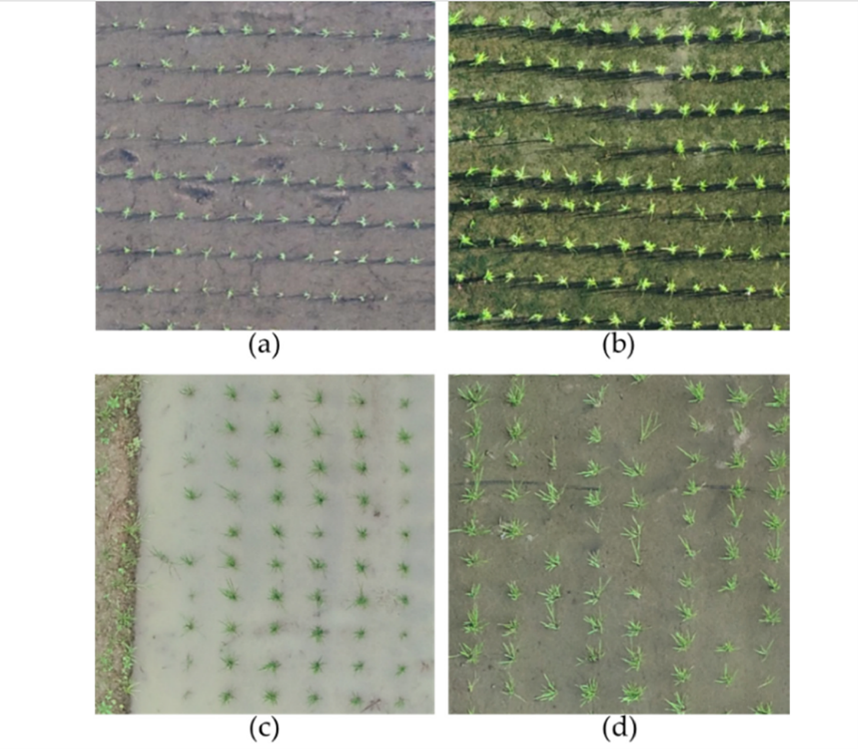} 
	\caption{An example of the test images on four datasets on (a) 7 August 2018, (b) 14 August (c) 12 August 2019 and (d) 20 August}
	\label{fig:generalization_analysis}
\end{figure}

\begin{figure}[ht]
	\centering
	\includegraphics[width=0.5\linewidth]{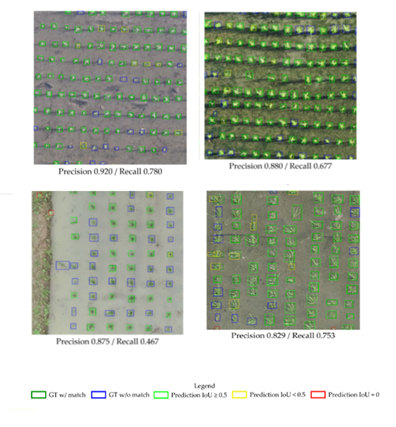} 
	\caption{An example of detection results with precision and recall metrics on four datasets}
	\label{fig:generalization_analysis}
\end{figure}

\section{Conclusion}

The detection of minute targets in high-resolution aerial imagery remains a formidable challenge in precision agriculture, compounded by the computational constraints of edge deployment and the visual complexity of heterogeneous field environments. This study establishes a robust framework for rice seedling detection by leveraging a Faster R-CNN architecture initialized via transfer learning. By integrating a semi-automated annotation pipeline, we effectively mitigated the data scarcity bottleneck, enabling the rapid curation of training datasets with minimal expert intervention.

Our empirical evaluation demonstrates that two-stage detectors can achieve high-precision localization even for objects with sparse feature representations. The model exhibited strong generalization across temporal domain shifts, maintaining Mean Average Precision (mAP) scores between $0.888$ and $0.986$ on three out of four test sets. These results validate the efficacy of transfer learning in adapting deep, generic feature extractors to the specific morphological constraints of rice seedlings. However, the distinct performance degradation observed on the final test set ($mAP=0.739$) highlights the sensitivity of current CNN architectures to severe environmental noise, specifically high-density algae growth and extreme lighting variations. This suggests that while the model is robust to minor temporal shifts, significant radiometric and textural deviations remain a barrier to fully autonomous deployment.

\paragraph{Future Work}
Future research will focus on two primary axes to address the identified limitations. First, to enhance robustness against environmental variance—specifically illumination changes, motion blur, and background clutter like algae—we will investigate unsupervised domain adaptation (UDA) and active learning techniques. These approaches aim to iteratively incorporate hard examples into the training manifold without necessitating exhaustive manual annotation, thereby bridging the domain gap observed in the August 20 dataset. Second, to facilitate deployment on resource-constrained UAV hardware, we will explore model compression strategies, including quantization and pruning, to optimize the trade-off between inference latency and detection accuracy for real-time edge computing.

\bibliographystyle{unsrtnat}
\bibliography{references}  

\end{document}